\documentclass[review]{elsarticle}
\usepackage{lineno,hyperref}
\usepackage[utf8]{inputenc}
\usepackage{graphicx}
\usepackage{amsmath}
\usepackage{amsfonts}
\usepackage{amssymb}
\usepackage{mathtools}
\usepackage[ruled,vlined]{algorithm2e}
\DeclareMathOperator*{\argmin}{argmin}
\DeclareMathOperator*{\argmax}{argmax}
\usepackage{enumitem}
\newtheorem{thm}{Theorem}

\newtheorem{lemma}{Lemma}

\modulolinenumbers[5]
\journal{Computational Statistics \& Data Analysis}
\begin{document}
	
	\begin{frontmatter}
		
		\title{Robust and Parallel Bayesian Model Selection}
		
		\author[ut]{Michael Minyi Zhang\corref{mycorrespondingauthor}}
		\ead{michael\_zhang@utexas.edu}
		\author[columbia]{Henry Lam}
		\ead{henry.lam@columbia.edu}
		\author[ndu]{Lizhen Lin}
		\ead{lizhen.lin@nd.edu}
		
		\cortext[mycorrespondingauthor]{Corresponding author. Address: Department of Statistics and Data Sciences. The University of Texas at Austin. 1 University Station G2550. Austin, TX 78712. Tel: (512) 232-0693. Fax: (512) 475-8297.}
		
		\address[ut]{The University of Texas at Austin, Austin, TX 78712, USA.}
		\address[columbia]{Columbia University, New York, NY 10027, USA.}
		\address[ndu]{University of Notre Dame, Notre Dame, IN 46556, USA.}
		\begin{abstract}
			Effective and accurate model selection is an important problem in modern data analysis. One of the major challenges is the computational burden required to handle large data sets that cannot be stored or processed on one machine. Another challenge one may encounter is the presence of outliers and contaminations that damage the inference quality. The parallel ``divide and conquer'' model selection strategy divides the observations of the full data set into roughly equal subsets and perform inference and model selection independently on each subset. After local subset inference, this method aggregates the posterior model probabilities or other model/variable selection criteria to obtain a final model by using the notion of geometric median. This approach leads to improved concentration in finding the ``correct" model and model parameters and also is provably robust to outliers and data contamination.


		\end{abstract}
		
		
	\end{frontmatter}
	
	\linenumbers

	\section{INTRODUCTION}
\label{sec:intro}
In many data modeling scenarios, many plausible models are available to fit to the data, each of which may result in drastically different predictions and conclusions. Being able to select the right model for inference is a crucial task. As our main example, we consider model selection for a normal linear model:
\begin{align}
Y &= X\beta + \epsilon, \hspace{.5em}\epsilon \sim N(0, \sigma^2 I)\label{linmod},
\end{align}
where $Y$ is an $N$ dimensional response vector, $X$ is an $N \times D$ dimensional design matrix and $\beta$ is a $D$ dimensional vector of regression parameters. Here the candidate models to be selected could refer to the sets of significant variables. In a Bayesian setting, we have a natural probabilistic evaluation of models through posterior model probabilities. Depending on the objectives of the data analysis, we may be interested in assessing the belief on which is the ``best'' model or obtaining predictions with minimum error.

Existing procedures to accomplish the aforementioned goals, however, will perform poorly under the presence of outliers and contaminations. In addition, Markov chain Monte Carlo (MCMC) algorithms for these methods do not scale to big data situations. The goal of this paper is to investigate a ``divide-and-conquer'' method that integrates with existing Bayesian model selection techniques, in a way that is robust to outliers and, moreover, allows us to perform Bayesian model selection in parallel.



Our ``divide-and-conquer" strategy is based on the ideas for robust inference using the notion of the geometric median \cite{minsker2015geometric}, especially the median posterior in the Bayesian context \cite{wang2014median,minsker2014scalable}. Previous work in this area has focused on the performance in parametric inference.
Our contribution in this paper is to demonstrate the effectiveness of these ideas in selecting the correct class of models on top of the parameters. In particular, we show that the model aggregated across different subsets (the ``divide") has improved concentration to the true model class compared to the one using the full data set. This concentration is in terms of the posterior model probabilities to the point mass assigned to the true model. The result also holds jointly with the concentration of the parameter estimates, and under the presence of outliers and hence demonstrates robustness. We carry out extensive numerical studies on  simulation data and a real data example to demonstrate the performance of our proposed approach.



\section{BAYESIAN MODEL SELECTION}
\label{sec:model_selection}

In Bayesian  model selection, we define the prior model probability $Pr(M_k)$ for each of the model $M_k$ ($k = 1, \ldots, K$) under consideration. For model $M_k$, we additionally have parameters $(\beta_k, \sigma_k^2)$ with prior $Pr(\beta_k,\sigma_k^2|M_k)$, which leads to a likelihood $Pr(Y|\beta_k,\sigma_k^2, M_k)$. Thus, the posterior model probability for model $M_k$, $ Pr(M_k |-) $, is proportional to
\begin{align*}
Pr(M_k) \int \!  Pr(Y| \beta_k,\sigma_k^2, M_k)Pr(\beta_k,\sigma_k^2|M_k) \, \mathrm{d}\beta_k\mathrm{d}\sigma_k^2.
\end{align*}
However, as noted in \cite{barbieri2004optimal}, choosing the model with the highest posterior model probability is not always the best option nor should one neglect the risk of model uncertainty. Instead of resorting to a single model for predicted values $\tilde{Y}$ (or some quantity of interest in general),  \cite{hoeting1999bayesian}  proposes to average over the model uncertainty with Bayesian model averaging (BMA) to obtain a posterior mean and variance of $\tilde{Y}$ at a covariate level $\tilde{X}$:
\begin{align*}
E[\tilde{Y} | \tilde{X},  Y] = \sum_{k=1}^{K}&E[\tilde{Y} | \tilde{X},  Y, M_k]Pr(M_k|\tilde{X},  Y),\\ \nonumber
Var(\tilde{Y} | \tilde{X}, Y) = \sum_{k=1}^{K}&Pr(M_k|X,  Y)\left( Var(\tilde{Y} | \tilde{X},  Y,  M_k) + \right.\\
&\left. E[\tilde{Y} | \tilde{X},  Y, M_k]^2 \right) - E[\tilde{Y} |X,  Y]^2.
\end{align*}
We will focus on BMA in our theoretical developments in this paper. Our numerical experiments, however, will show that our divide-and-conquer strategy is also effective in applying on other model selection methods. 

The first alternative to BMA is the median probability model, which can be shown to be optimal if we must choose one model for prediction \cite{barbieri2004optimal}. In this approach, we define the posterior inclusion probability of each predictor $x_d$ ($d=1,\ldots, D$) as the sum of posterior model probabilities of the models that include predictor $x_d$, namely $p_d = \sum_{k : x_d \in M_k} Pr(M_k |X,  Y)$. The median probability model is  the model that includes the predictors $x_d$ if $p_d \geq 1/2$.

Second, using the maximum value of the likelihood for each model $Pr(Y | \hat\beta_k,\hat\sigma_k^2, M_k)$, where $(\hat\beta_k,\hat\sigma_k^2)$ is the maximum likelihood estimate of $(\beta_k,\sigma_k^2)$, we can perform penalized model selection through the Akaike information criterion (AIC) \cite{akaike1974new} or  the Bayesian information criterion (BIC) \cite{schwarz1978estimating} by selecting the model with the lowest information criterion:
\begin{align*}
\text{AIC} = -2 \log Pr(Y | \hat\beta_k,\hat\sigma_k^2, M_k) + 2(D+1),\\
\text{BIC} = -2 \log Pr(Y | \hat\beta_k,\hat\sigma_k^2, M_k) + (D+1)\log N.
\end{align*}

The final model selection technique we will consider is stochastic variable selection through the spike and slab model \cite{george1993variable}, which allows for variable shrinkage under high-dimensional models.
For the purposes of this paper, we will use the rescaled spike and slab model \cite{ishwaran2005spike}.  To perform posterior inference in this model, we first define $Y^{\prime} = \sqrt{\frac{N}{\hat{\sigma}^2}}Y $ where $\hat{\sigma}^2$ is the unbiased estimate of $\sigma^2$ under the full model and let  $\nu_0 > 0$  be some small number. The model is defined to be the following mixture model:
\begin{align*}
\begin{split}
Y^{\prime} &\sim N(X\beta, N\sigma^2I), \; \beta_{SS_d} \sim N(0, J_d\tau_d^2),\\
\sigma^{-2}_{ss} &\sim \text{Gamma}(a,b),\; J_d \sim (1-w)\delta_{J_d}(\nu_0) + w\delta_{J_d}(1),\\
\tau_d^{-2} &\sim \text{Gamma}(a_{\tau}, b_{\tau}),\;w \sim \text{Uniform}(0,1).
\end{split}
\end{align*} 
	\section{DIVIDE-AND-CONQUER AND ROBUST BAYESIAN MODEL SELECTION}
\label{sec:method}

In our robust model selection strategy, we divide $N$ observations  into $R$ subsets of roughly equal sample size. Then inference, model selection and prediction is performed for the linear model $Y_{(j)} = X_{(j)}\beta + \epsilon_{(j)}$ independently across $j=1,\ldots , R$ subsets using the existing Bayesian model selection procedures, which are then combined to form a final model or a combined prediction value.

Given linear model \eqref{linmod}, we first define the following priors on a normal likelihood with response variable $Y$ and $D$-dimensional predictor $X$. The $N$ observations  are divided into $R$ subsets with $s$ observations within each subset. One has,
\begin{align*}
Pr\left( \sigma^{-2}_{(j)} \right) &= \text{Gamma}(a,b),\\
Pr(\beta_{(j)} | \sigma^2_{(j)}) &= N(\beta_0, \sigma^2_{(j)} \Sigma_0).
\end{align*}
To compensate for the data division, we raise the likelihood of the divided data $Pr(Y_{(j)}| X_{(j)}, \beta, \sigma^{2} )$ to the $R$-th power and adjust the normalizing constant accordingly so that the likelihood for $Y_{j}$ is:
\begin{align*}
\left( \frac{R}{2\pi \sigma^2_{(j)}} \right)^{N/2} \exp\left\{ \frac{-R}{2\sigma^2} \left(Y_{(j)} - X_{(j)}\beta_{(j)} \right)^T \left(Y_{(j)} - X_{(j)}\beta_{(j)} \right) \right\}.
\end{align*}
The  intuition and motivation for raising the subset likelihood to $R$-th power is to adjust  the potentially inflated variance of the subset posterior distribution.
Exploiting conjugacy, we obtain the full conditionals for data subset $ j = 1, \ldots , R $:
\begin{align*}
Pr(\beta_{(j)} | - ) &\nonumber = N\left( \mu_\beta, \sigma^2 \Sigma_\beta \right),\\
\mu_\beta &= \Sigma_\beta\left( \beta_0 \Sigma_0^{-1} + R  X_{(j)}^T Y_{(j)} \right),\\
\Sigma_\beta &= \left( \Sigma_0^{-1} + R X_{(j)}^{T} X_{(j)} \right)^{-1},\\ \nonumber
Pr\left( \sigma^{-2}_{(j)} | - \right) & = \text{Gamma}\left( a^{\prime}, b^{\prime}\right),\\ \nonumber
a^{\prime} &= a + \frac{N+D}{2},\\
b^{\prime} &= b + \frac{R}{2} \epsilon^T \epsilon + \frac{1}{2}\left( \beta_{(j)} - \beta_0 \right)^T \Sigma_0^{-1}\left( \beta_{(j)} - \beta_0 \right), \\
\epsilon &= \left(Y_{(j)} - X_{(j)}\beta_{(j)}\right).
\end{align*}
Let $\Sigma_X =  I + RX_{(j)}\Sigma_0 X_{(j)}^T$, then integrating out the parameters gives us the following marginal distribution $Pr(Y_{(j)}|X_{(j)})$:
\begin{align*}
  \frac{ \left( \frac{R}{2\pi} \right)^{ \frac{N}{2} } b^a  \Gamma(a+ \frac{N}{2})\left|\Sigma_X \right|^{-\frac{1}{2}} /\, \Gamma(a) }{\left(b+ \frac{R}{2}\left(Y_{(j)} - X_{(j)}\beta_0 \right)^T \Sigma_X^{-1} \left(Y_{(j)} - X_{(j)}\beta_0 \right) \right)^{a + \frac{N}{2} }}.
\end{align*}

For distributed AIC and BIC model evaluation, we raise the likelihood term of the AIC and BIC formula to the power of $R$:
\begin{align*}
\text{AIC}_R = -2R \log Pr(Y_{(j)} | \hat\beta_k,\hat\sigma_k^2, M_k) + 2(D+1), \\
\text{BIC}_R = -2R \log Pr(Y_{(j)} | \hat\beta_k,\hat\sigma_k^2, M_k) + (D+1)\log N.
\end{align*}

In applying our procedure with the spike and slab prior, we derived the full Gibbs sampler for our procedure. For posterior inference in the spike and slab model, let $\Delta = \text{diag}\left\{ J_1\tau_{1}^{2}, \ldots , J_D\tau_{D}^{2} \right\}$,  we can perform Gibbs sampling by drawing from the following posteriors:
\begin{align*}
Pr(\beta_{SS(j)} | - ) & = N(\mu_{\beta_{SS}}, \Sigma_{\beta_{SS}}),\\
\Sigma_{\beta_{SS}} &= \left( \Delta^{-1} + \frac{R}{N\sigma^{-2}_{SS(j)}} X_{(j)}^TX_{(j)} \right)^{-1},\\ 
\mu_{\beta_{SS}} &= \Sigma_{\beta_{SS}} \left( \frac{R}{N\sigma^{-2}_{SS(j)}} X_{(j)}^TY_{(j)} \right),\\
Pr\left( \sigma^{-2}_{SS(j)} | - \right) & = \text{Gamma}\left( a^{\prime}_{SS}, b^{\prime}_{SS} \right),\\
a^{\prime}_{SS} &= a + \frac{N}{2},\\
b^{\prime}_{SS} &= b + \frac{R}{2N}\left(Y_{(j)} - X_{(j)}\beta_{SS(j)} \right)^T \left(Y_{(j)} - X_{(j)}\beta_{SS(j)} \right),\\
\begin{split}
Pr\left( J_d | - \right) &\propto w_{d1} \delta_{J_d}(\nu_0) + w_{d2}\delta_{J_d}(1),\\
w_{d1} &= (1-w)\nu_0^{-1/2}\exp\left\{ -\frac{\beta_{SS(j)d}^2}{2\nu_0\tau^{2}_{d}} \right\},\\
w_{d2} &= w \exp\left\{ -\frac{\beta^{2}_{SS(j)d}}{2\tau^{2}_{d}} \right\},
\end{split}\\
Pr\left( \tau^{-2}_{d} | - \right) & = \text{Gamma}\left( a_{\tau} + \frac{1}{2}, b_{\tau} + \frac{\beta^{2}_{SS(j)d}}{2J_{d}} \right),\\
Pr(w | -) & = \text{Beta}\left( 1+ \left| \left\{ d : J_d = 1 \right\} \right| , 1+ \left| \left\{ d : J_d = \nu_0 \right\} \right| \right).
\end{align*}


Once inference is built on each subset, the key step is to aggregate the subset models (or estimates) together into a final model (or estimate). To aggregate our results, we collect the $R$ number of subset models or estimates and find the geometric median between these $R$ elements. The geometric median for a set of elements $\{x_1,\ldots,x_R\}$ valued on a Hilbert space $\mathbb H$, is defined as
\begin{align}
\begin{split}
x_*&=\text{med}_g(x_1,\ldots,x_R)=\text{argmin}_{y\in\mathbb H}\sum_{j=1}^R\|y-x_j\|,\label{geometric median definition}
\end{split}
\end{align}
where $\|\cdot\|$ is the norm associated with the inner product in $\mathbb H$ \cite{minsker2014scalable}. The solution can generally be effectively approximated using the Weiszfeld algorithm \cite{wes37}.

For instance, in the case of aggregating the posterior model probabilities across $R$ subsets of data, the geometric median operates on the space of posterior distributions and the geometric median posterior model probability, $ Pr_*(M_k|X, Y) $, is defined as:
\begin{align}
\argmin_{P \in \Pi_K} \sum_{j=1}^{R} \left|\left| P - Pr(M_k|X_{(j)}, Y_{(j)}) \right|\right|\label{model aggregation},
\end{align}
where $Pr(M_k|X_{(j)}, Y_{(j)})$ is the posterior model probabilities for subset $j$, and $\Pi_K$ denotes the space of distributions on $K$ support points. The metric $\|\cdot\|$ here can be taken as the Euclidean metric, or an integral probability metric (IPM) defined as $||P - Q|| = \sup_{f \in \mathcal{F}} \left| \int \! f(x) \, \mathrm{d}(P-Q)(x) \right|$ for some class of functions $\mathcal F$ \cite{sriperumbudur2010hilbert,sriperumbudur2012empirical}.

For the model selection techniques discussed earlier (AIC, BIC, and the median model selection), we can choose a final model in two ways: One, we can select the best model locally on each subset, use it for prediction, and then aggregate the results (estimate combination). Or two, we can take the median of the model selection criteria and choose that particular model on each subset and then aggregate the results to get a final model (model combination).

However, in Bayesian model averaging and spike and slab modeling we do not choose a final model. We can still perform model or estimate combination by aggregating the posterior model probabilities. We consider both model and estimate combinations in our experiments and show that they yield similar results in our experimental settings.

\begin{algorithm}
\caption{Algorithm for robust model selection in the case of BMA.}
\label{alg:model_select}
\For{$j \in \left\{ 1,\ldots , R \right\} $}{
		Raise likelihood to $R$-th power\\
		Compute inference for $P(\theta | M_k, X_{(j)},Y_{(j)})$ for $k=1,\ldots,K$\\
		Draw predictive values from predictive posterior $P(\tilde{Y} | M_k, X_{(j)},Y_{(j)})$ for $k=1,\ldots,K$\\
Calculate posterior model probabilities $\{P(M_k | X_{(j)},Y_{(j)})\}_{k=1,\ldots,K}$\\
}
	Calculate geometric median of posterior model probabilities over the subsets using \eqref{model aggregation}.\\
Approximate geometric medians of posterior parameter probabilities or predictive values given individual models over the subsets using \eqref{geometric median definition}.\\
			Obtain BMA estimate: $E[\tilde Y | Y, X] = \sum_{k=1}^{K}E_*[\tilde Y | X, Y, M_k]Pr_*(M_k|X, Y)$
	
\end{algorithm}

	\section{IMPROVED CONCENTRATION AND ROBUSTNESS}
\label{sec:theory}
In this section we provide theoretical justification on the robustness in the divide-and-conquer strategy. In particular, we focus on BMA. Additionally, we show that the aggregated model class from our strategy concentrates faster, in terms of posterior model probabilities, to the correct class compared to using the whole data set at once. This concentration result can be joint with parameter estimation, and also applies in a way that exhibits robustness against outliers. Note that we do not  raise the subset likelihood to $R$-th power in our current theoretical analysis, but the results can be generalized by imposing slightly stronger entropy conditions on the model.

Let $\mathcal S$ be the domain of $\theta=(M_k,\beta,\sigma^2)$, our set of model indices and parameters. Let $\theta_0$ be the true data generating parameter, and let $(X_1,Y_1)$ be a generic data point. Let $p_0(y|x):=p(y|x,\theta_0)$ be the true conditional density of $Y_1$ given $X_1$, and $p_0(x)$ be the true density of the covariates $X_1$. We denote $p_\theta(y|x):=p(y|x,\theta)$. Let $P_\theta$ be the distribution defined by $p_0(x)\times p_\theta(y|x)$ and $P_0$ is the true distribution $p_0(x)\times p_0(y|x)$. For convenience, we  denote $P_0f=P_0f(X_1,Y_1)=E_{p_0}[f(X_1,Y_1)]$ where $E_{p_0}[\cdot]$ is the expectation under $p_0(y|x)\times p_0(x)$. We denote $P_0^N$ as the true probability measure taken on the data $(X,Y)$ of size $N$ and  $P_0^Nf=E_{P_0^N}[f(X,Y)]$. Lastly, we denote $\mathcal D(\epsilon,\mathcal P,d)$ as the $\epsilon$-packing number of a set of probability measures $\mathcal P$ under the metric $d$, which is the maximal number of points in $\mathcal P$ such that the distance between any pair is at least $\epsilon$. We implicitly assume here that $\mathcal P$ is separable. The following Theorem \ref{thm plain} follows from a modification of Theorem 2.1 in \cite{ghosal2000}:
\begin{thm}
Assume that there is a sequence $\varepsilon_N$ such that $\varepsilon_N\to0$ and $N\varepsilon_N^2\to\infty$ as $N\to\infty$, a constant $C$, and a set $\mathcal S_N\in\mathcal S$ so that
\begin{enumerate}[leftmargin=*]
\item $\log\mathcal D(\varepsilon_N/2,\mathcal P_{\mathcal S_N},d_H)\leq N\varepsilon_N^2.$\label{assumption1}
\item $Pr(\mathcal S\setminus\mathcal S_N)\leq e^{-N\varepsilon_N^2(C+4)}.$\label{assumption2}
\item $Pr\left(\theta:-P_0\log\frac{p_\theta(Y_1|X_1)}{p_0(Y_1|X_1)}\leq\varepsilon_N^2,\ \right. \\
\left. \qquad  P_0\left(\frac{p_\theta(Y_1|X_1)}{p_0(Y_1|X_1)}\right)^2\leq\varepsilon_N^2\right)\geq e^{-N\varepsilon_N^2C}.$\label{assumption3}
\end{enumerate}
where $\mathcal P_{\mathcal S_N}=\{p_0(x)\times p_\theta(y|x):\theta\in\mathcal S_N\}$ and $d_H$ is the Hellinger distance. Then we have
\begin{align}
\begin{split}
P_0^N&\left(Pr(\theta:d_H(P_\theta,P_0)>T\varepsilon_N^2|X,Y)>\delta\right)\leq \\
&\frac{1}{C^2N\varepsilon_N^2\delta}+\frac{2e^{-LN\varepsilon_N^2}}{\delta}+\frac{2e^{-2N\varepsilon_N^2}}{\delta},\label{main plain}
\end{split}
\end{align}
for any $0<\delta<1$ and sufficiently large $T>0$ such that $LT^2\geq C+4$ and $LT^2-1>L$, where $L$ is a universal constant.\label{thm plain}
\end{thm}

The proof of Theorem \ref{thm plain} is in the Appendix. As noted by \cite{ghosal2000}, the important assumptions are Assumptions \ref{assumption1} and \ref{assumption3}. Essentially, Assumption \ref{assumption1} constrains the size of the parameter domain $\mathcal S$ to be not too big, whereas Assumption \ref{assumption3} ensures sufficient mass of the prior  on a neighborhood of the true parameter. The concentration result \eqref{main plain} states that the posterior distribution of $\theta$ is close to the true $\theta_0$ with high probability, where the closeness is measured in terms of the Hellinger distance between the likelihoods. Note that the RHS of \eqref{main plain} consists of three terms. The dominant term is the power-law decay in $N\varepsilon_N^2$. The other two exponential decay terms result from technical arguments in the existence of tests that sufficiently distinguish between distributions \cite{birge1983approximation,le2012asymptotic}.

Next we describe the concentration behavior of BMA. We focus on the situations where all the candidate models are non-nested, i.e. only one model contains distributions that are arbitrarily close to the truth. Without loss of generality, we let $M_1$ be the true model.

\begin{thm}[BMA of Non-Nested Models]
Suppose the assumptions in Theorem \ref{thm plain} hold. Also assume that, for sufficiently small $\epsilon>0$, $d(P_\theta,P_0)>\epsilon$ for any $\theta\in\mathcal S_{-1}:=\{(M_k,\beta,\sigma^2):k\neq1\}$. Let $L$ be the same universal constant arising in Theorem \ref{thm plain}. We have
\begin{enumerate}[leftmargin=*]
\item For any given $0<\delta<1$,
\begin{align}
\begin{split}
P_0^N&\left(Pr(M_1|X,Y)<1-\delta\right)\leq \\
&\frac{1}{C^2N\varepsilon_N^2\delta}+\frac{2e^{-LN\varepsilon_N^2}}{\delta}+\frac{2e^{-2N\varepsilon_N^2}}{\delta},
\end{split}
\label{interim9}
\end{align}\label{plain}
for sufficiently large $N$.
\item For any given $0<\delta<1$,
\begin{align}
\begin{split}
P_0^N&\left(d_E(Pr(M_k|X,Y),\mathbf e_1)>\delta\right)\leq\\
&\frac{\sqrt 2}{C^2N\varepsilon_N^2\delta}+\frac{2\sqrt 2e^{-LN\varepsilon_N^2}}{\delta}+\frac{2\sqrt 2e^{-2N\varepsilon_N^2}}{\delta},
\end{split}
\label{Euclidean bound}
\end{align}
for sufficiently large $N$, where $d_E$ is the Euclidean distance, and $\mathbf e_1$ is the point mass on $M_1$.\label{Euclidean}
\item For any $0<\delta<\sqrt{(\sqrt 2-1)^2+1}/2$,
\begin{align}
\begin{split}
P_0^N&\left(d_H(Pr(M_k|X,Y),\mathbf e_1)>\delta\right)\leq \\
&\frac{(\sqrt 2-1)^2+1}{\sqrt 2C^2N\varepsilon_N^2\delta^2}+\frac{((\sqrt 2-1)^2+1)e^{-LN\varepsilon_N^2}}{\delta^2}+\\
&\frac{((\sqrt 2-1)^2+1)e^{-2N\varepsilon_N^2}}{\delta^2},
\end{split}
\label{Hellinger bound}
\end{align}
for sufficiently large $N$.
\label{Hellinger}
\end{enumerate}
\label{thm model selection}
\end{thm}

\noindent\textbf{Proof of Theorem \ref{thm model selection}.}

\emph{Proof of \ref{plain}.} Consider large enough $N$ and fix a sufficiently large $T>0$. We have
\begin{eqnarray}
&&Pr(\theta:d(P_\theta,P_0)\leq T\varepsilon_N^2|X,Y)\notag\\
&=&E_{Pr}\left[Pr(\theta:d(P_\theta,P_0)\leq T\varepsilon_N^2|M_k,X,Y)|X,Y\right]{},\\
&&{}\text{\ \ where $E_{Pr}[\cdot|X,Y]$ denotes the posterior expectation}{}\notag\\
&&{}\text{\ \ and $Pr(\cdot|M_k,X,Y)$ denotes the posterior distribution given model $M_k$}\notag\\
&=&Pr(M_1|X,Y)Pr(\theta:d(P_\theta,P_0)\leq T\varepsilon_N^2|M_1,X,Y),\label{interim14}
\end{eqnarray}
by the condition that $d(P_\theta,P_0)>T\varepsilon_N^2$ for any $\theta\in\mathcal S_{-1}$ and any $T>0$ eventually. Hence
\begin{equation}
Pr(\theta:d(P_\theta,P_0)\leq T\varepsilon_N^2|X,Y)\geq1-\delta,\label{interim7}
\end{equation}
implies
\begin{equation}
Pr(M_1|X,Y)\geq1-\delta\label{interim8}.
\end{equation}
The result then follows from Theorem \ref{thm plain}, which implies that \eqref{interim7} occurs with probability at least
$$1-\left(\frac{1}{C^2N\varepsilon_N^2\delta}+\frac{2e^{-LN\varepsilon_N^2}}{\delta}+\frac{2e^{-2N\varepsilon_N^2}}{\delta}\right),$$

\emph{Proof of \ref{Euclidean}.}
Note that \eqref{interim8} implies
\begin{equation}
d_E(Pr(M_k|X,Y),\mathbf e_1)=\sqrt{(1-Pr(M_1|X,Y))^2+\sum_{k\neq1}Pr(M_k|X,Y)^2}\leq\sqrt2\delta,\label{interim10}
\end{equation}
since $(1-Pr(M_1|X,Y))^2\leq\delta^2$ and
$(\delta,0,\ldots,0)$ is an optimizer of the optimization
$$\max\sum_{i=2}^Kx_i^2\text{\ \ subject to\ \ }\sum_{i=2}^Kx_i\leq\delta.$$
Hence \eqref{interim9} and \eqref{interim10} together imply
$$P_0^N\left(d_E(Pr(M_k|X,Y),\mathbf e_1)\geq\sqrt2\delta\right)\leq\frac{1}{C^2N\varepsilon_N^2\delta}+\frac{2e^{-LN\varepsilon_N^2}}{\delta}+\frac{2e^{-2N\varepsilon_N^2}}{\delta}.$$
By redefining $\tilde\delta=\sqrt2\delta$, we get \eqref{Euclidean bound}.

\emph{Proof of \ref{Hellinger}.}
Note that \eqref{interim8} implies
\begin{align}
d_H(Pr(M_k|X,Y),\mathbf e_1)&=\sqrt{\frac{1}{2}\left((\sqrt{1-Pr(M_1|X,Y)})^2+\sum_{k\neq1}Pr(M_k|X,Y)\right)}\notag\\
&\leq\sqrt{\frac{1}{2}\left((1-\sqrt{1-\delta})^2+\delta\right)},\label{interim11}
\end{align}
since $x_i=\delta/(k-1)$ for all $i\neq0$ gives the optimizer of the optimization
$$\max\sum_{i\neq0}\sqrt{x_i}\text{\ \ subject to\ \ }\sum_{i\neq0}x_i\leq\delta.$$
Hence \eqref{interim9} and \eqref{interim11} together imply
\begin{align}
\begin{split}
P_0^N\left(d_H(Pr(M_k|X,Y),\mathbf e_1)>\sqrt{\frac{1}{2}\left((1-\sqrt{1-\delta})^2+\delta\right)}\right)\leq\\
\frac{1}{C^2N\varepsilon_N^2\delta}+\frac{2e^{-LN\varepsilon_N^2}}{\delta}+\frac{2e^{-2N\varepsilon_N^2}}{\delta}\label{interim12}.
\end{split}
\end{align}
Note that $(1-\sqrt{1-\delta})^2$ is a convex function in $\delta$ for $0<\delta<1$ and is equal to 0 at $\delta=0$. Thus $(1-\sqrt{1-\delta})^2\leq(\sqrt 2-1)^2\delta$ for $0<\delta<1/2$, where $(\sqrt 2-1)^2$ is the slope of the line between $(0,0)$ and $(1/2,(1-\sqrt{1-1/2})^2$. Hence, for $0<\delta<1/2$, we have
$$\sqrt{\frac{1}{2}\left((1-\sqrt{1-\delta})^2+\delta\right)}\leq\sqrt{((\sqrt 2-1)^2+1)\frac{\delta}{2}}.$$
Combining with \eqref{interim12}, we have
\begin{align}
\begin{split}
P_0^N\left(d_H(Pr(M_k|X,Y),\mathbf e_1)>\sqrt{((\sqrt 2-1)^2+1)\frac{\delta}{2}}\right)\leq\\
\frac{1}{C^2N\varepsilon_N^2\delta}+\frac{2e^{-LN\varepsilon_N^2}}{\delta}+\frac{2e^{-2N\varepsilon_N^2}}{\delta}.
\end{split}
\end{align}
By redefining $\tilde\delta=\sqrt{((\sqrt 2-1)^2+1)\delta/2}$, we get \eqref{Hellinger bound}.\hfill\qed

Note that the assumption $d(P_\theta,P_0)>\epsilon$ for any $\theta\in\mathcal S_{-1}$ and sufficiently small $\epsilon$ is a manifestation of the non-nested model situation, asserting that only one model is ``correct". Result \ref{plain} is a concentration on the posterior probability of picking the correct model to be close to 1.

Result \ref{Euclidean} translates this in terms of the Euclidean distance between the model posterior probability and the point mass on the correct model. Result \ref{Hellinger} is an alternative using the Hellinger distance.
Note that the concentration bound for Hellinger distance \eqref{Hellinger bound} is inferior to that for Euclidean distance \eqref{Euclidean bound} for small $\delta$ since $\delta^2$ instead of $\delta$ shows up in the RHS of \eqref{Hellinger bound}. This is because in our proof, the function $\sqrt{(1-\sqrt{1-\delta})^2+\delta}$ that appears in \eqref{interim12} has derivative $1/(2\sqrt{(1-\delta)((1-\sqrt{1-\delta})^2+\delta)})$ which is $\infty$ at $\delta=0$, and thus no linearization is available when $\delta$ is close to 0.

Theorem \ref{thm model selection} can be modified to handle the case where multiple models contain the truth. In particular, the expression inside the probability in \eqref{interim9} becomes
$$\sum_{r\in\mathcal M}Pr(M_r|X,Y)<1-\delta,$$
where $\mathcal M$ is the collection of all $r$ such that $M_r$ contains the true model. In \eqref{Euclidean bound} and \eqref{Hellinger bound}, the use of $\mathbf e_1$ is replaced by an existence of some probability vector (dependent on $N$) supported on the indices in $\mathcal M_r$. In other words, one now allows comparing with an arbitrary allocation of probability masses to all true models in the concentration bound. These modifications can be seen by following the arguments in the proof of Theorem \ref{thm model selection}. Specifically, \eqref{interim14} would be modified as
$$\sum_{r\in\mathcal M}Pr(M_r|X,Y)Pr(\theta:d(P_\theta,P_0)\leq T\varepsilon_N^2|M_r,X,Y).$$
Then \eqref{interim7} would imply a modified version of \eqref{interim8}, namely
$$\sum_{r\in\mathcal M}Pr(M_r|X,Y)\geq1-\delta,$$
giving the claimed modification for \eqref{interim9}. Then, following \eqref{interim10}, we could find a probability vector to make all $(1-Pr(M_r|X,Y))^2$ terms vanish except one, which is in turn bounded by $\delta^2$. This gives the claimed modifications for \eqref{Euclidean bound} and \eqref{Hellinger bound}.




The following result states how a divide-and-conquer strategy can improve the concentration rate of the posterior model probabilities towards the correct model:
\begin{thm}[Concentration Improvement]
	Suppose the assumptions in Theorem \ref{thm model selection} hold. Let $s=N/R$, and
	$q=\frac{\sqrt 2}{C^2s\varepsilon_s^2\delta}+\frac{2\sqrt 2e^{-Ls\varepsilon_s^2}}{\delta}+\frac{2\sqrt 2e^{-2s\varepsilon_s^2}}{\delta}$.
	For sufficiently large $s$, letting $\alpha,\nu$ be constants such that $0<q<\alpha<1/2$ and $0\leq\nu<(\alpha-q)/(1-q)$, we have:
	\begin{enumerate}[leftmargin=*]
		\item $Pr_*(M_k|X,Y)$, the geometric median under $d_E$ of $\{Pr(M_k|(X_{(j)},Y_{(j)}))\}_{j=1,\ldots,R}$, satisfies
		\begin{align}
		\begin{split}
		P_0^N&\left(d_E(Pr_*(M_k|X,Y),\mathbf e_1)>C_\alpha\delta\right)\leq\\
		&\left(e^{(1-\nu)\psi(\frac{\alpha-\nu}{1-\nu},q)}\right)^{-R},
		\end{split}\label{geometric median Euclidean bound eqn}
		\end{align}\label{geometric median Euclidean bound}
		where $C_\alpha=(1-\alpha)\sqrt{1/(1-2\alpha)}$, and $\psi(\alpha,q)=(1-\alpha)\log\frac{1-\alpha}{1-q}+\alpha\log\frac{\alpha}{q}$.
		\label{geometric median Euclidean}
		\item Let $K$ be the number of model classes, then:
		\begin{align}
		\begin{split}
		P_0^N&\left(Pr_*(M_1|X,Y)<1-C_\alpha\delta\sqrt{\frac{K-1}{K}}\right)\leq \\
		&\left(e^{(1-\nu)\psi(\frac{\alpha-\nu}{1-\nu},q)}\right)^{-R}.
		\end{split}\label{geometric median plain eqn}
		\end{align}\label{geometric median plain}
		\item Suppose in addition that, for any $P_{\theta^1},P_{\theta^2}$ such that $\theta^i=(M_1,\beta^i,(\sigma^2)^i)$ for $i=1,2$, we have
		\begin{equation}
		d_H(P_{\theta^1},P_{\theta^2})\geq\tilde C\rho_k(\theta^1,\theta^2)^\gamma,\label{IPM}
		\end{equation}
where $\rho_k(\theta^1,\theta^2)=\|k(\cdot,\theta^1)-k(\cdot,\theta^2)\|_{\mathbb H}$, with $k$ being a characteristic kernel defined on the space $\{\theta=(M_1,\cdot,\cdot)\}$ and $\mathbb H$ is the corresponding reproducing kernel Hilbert space (RKHS), and $\tilde C>0$ and $\gamma>0$ are constants. Moreover, assume that there is a universal constant $\tilde K$ such that $e^{-\tilde Ks\varepsilon_s^2/2}\leq\varepsilon_s$ for all $s$, and we choose $\varepsilon_s$ such that $\tilde q=\frac{1}{Cs\varepsilon_s^2}+4e^{-\tilde Ks\varepsilon_s^2/2}<1/2$. Then
\begin{align*}
		\begin{split}
		P_0^N\left(Pr_*(M_1|X,Y)>1-C_\alpha\delta\sqrt{\frac{K-1}{K}}, \right. \\
		\left. \|Pr_*(\theta|M_1,X,Y)-\delta_0\|_{\mathcal F_k}\leq C_\alpha\tilde T\epsilon_s^{1/\gamma}\right)\\
\geq1-\left(e^{(1-\nu)\psi\left(\frac{\alpha-\nu}{1-\nu},q\right)}\right)^{-R}-\left(e^{\psi\left(\alpha,q\right)}\right)^{-R},
		\end{split}
		\end{align*}
where $\|\cdot\|_{\mathcal F_k}$ is defined as $\|P-Q\|_{\mathcal F_k}=\|\int k(x,\cdot)d(P-Q)(x)\|_{\mathbb H}$, $\tilde T>0$ is a sufficiently large constant, $Pr_*(\theta|M_1,X,Y)$ is the geometric median of $\{Pr(\theta|M_1,(X_{(j)},Y_{(j)}))\}_{j=1,\ldots,R}$ under the $\|\cdot\|_{\mathcal F_k}$-norm, and $\delta_0$ is the delta measure at the true parameter.\label{geometric median joint}

	\end{enumerate}\label{thm main}
\end{thm}

The significance of Theorem \ref{thm main} is the improvement of the concentration from power-law decay in Theorem \ref{thm model selection} to exponential decay, as the number of subsets grows. Such type of results is known in the case of parameter estimation (e.g., \cite{wang2014median,minsker2014scalable}). Theorem \ref{thm main} generalizes to the case of model selection.
 Results \ref{geometric median Euclidean} and \ref{geometric median plain} describe the exponential concentration for the model posteriors to the correct model, while Result \ref{geometric median joint} states the joint concentration in both the model posterior and the parameter posterior given the correct model, when one adopts a second layer of divide-and-conquer on the parameter posterior conditional on each individual candidate model. Result \ref{geometric median joint} in particular combines with the parameter concentration result in \cite{minsker2014scalable}.

Note that we have taken a hybrid viewpoint here that we assume a ``correct" model and parameters in a frequentist sense. Under this view, a posterior probability more concentrated towards the truth is more desirable. This constitutes our main claim that the divide-and-conquer strategy is attractive. This view has been used in existing work like \cite{wang2014median,minsker2014scalable}.

Finally, the following theorem highlights that the concentration improvement still holds even if the data are contaminated to a certain extent:
\begin{thm}[Robustness to Outliers]
Using the notation in Theorem \ref{thm main}, but assume instead that, for $j$ where $1\leq j\leq\lfloor(1-\nu)R\rfloor+1$,
\begin{align*}
\begin{split}
P_0^s&\left(d_E(Pr(M_k|X_{(j)},Y_{(j)}),\mathbf e_1)>\delta\right)\leq\\
&\frac{\sqrt 2}{C^2s\varepsilon_s^2\delta}+\frac{2\sqrt 2e^{-Ls\varepsilon_s^2}}{\delta}+\frac{2\sqrt 2e^{-2s\varepsilon_s^2}}{\delta},
\end{split}
\end{align*}
the conclusion of Theorem \ref{thm main} still holds.\label{thm extension}
\end{thm}
Theorem \ref{thm extension} stipulates that when a small number of subsets are contaminated by arbitrary nature, the geometric median approach still retains the same exponential concentration. 

\noindent\textbf{Proofs of Theorems \ref{thm main} and \ref{thm extension}.}

The proofs of both theorems rely on a key theorem on geometric median in \cite{minsker2015geometric}, restated in the Appendix. We focus on Theorem \ref{thm main}, as the proof for Theorem \ref{thm extension} is a straightforward modification in light of Theorem \ref{thm geometric median}.

\emph{Proof of \ref{geometric median Euclidean}.}
Immediate by noting that
$$P_0^s\left(d_E(Pr(M_k|X_{(j)},Y_{(j)}),\mathbf e_1)>\delta\right)\leq q,$$
for all $j=1,\ldots,R$, and applying Theorem \ref{thm geometric median}.

\emph{Proof of \ref{geometric median plain}.}
Note that
\begin{equation}
d_E(Pr_*(M_k|X,Y),\mathbf e_1)\geq(1-Pr_*(M_1|X,Y))\sqrt{\frac{K}{K-1}}\label{interim13}.
\end{equation}
To see this, let $a=Pr_*(M_1|X,Y)$. We have $$d_E(Pr_*(M_k|X,Y),\mathbf e_1)=\sqrt{(1-a)^2+\sum_{i=2}^Kx_i^2},$$ where $x_i$'s satisfy $\sum_{i=2}^Kx_i=1-a$. Since $(1-a)/(K-1)$ is the optimizer of the optimization
$$\min\sum_{i=2}^Kx_i^2\text{\ \ subject to\ \ }\sum_{i=2}^Kx_i=1-a,$$
we get $\sqrt{(1-a)^2+\sum_{i=2}^Kx_i^2}\geq(1-a)\sqrt{K/(K-1)}$.

Hence \eqref{geometric median Euclidean bound eqn} and \eqref{interim13} together give
$$P_0^N\left(Pr_*(M_1|X,Y)<1-C_\alpha\delta\sqrt{\frac{K-1}{K}}\right)\leq\left(e^{(1-\nu)\psi(\frac{\alpha-\nu}{1-\nu},q)}\right)^{-R}.$$

\emph{Proof of \ref{geometric median joint}.}
Under the additional assumptions, we can invoke Corollary 3.5 in \cite{minsker2014scalable} to obtain that
$$P_0^N\left(\|Pr_*(\theta|M_1,X,Y)-\delta_0\|_{\mathcal F_k}>C_\alpha\tilde T\epsilon_s^{1/\gamma}\right)\leq\left(e^{\psi\left(\alpha,q\right)}\right)^{-R}.$$
The result follows from applying a union bound and together with \eqref{geometric median plain eqn}.
\hfill\qed

	\section{SIMULATIONS AND DATA ANALYSIS}\label{sec:sim}
For the BMA, AIC, BIC and median probability model tests, we generate data from a model $Y=X\beta + \epsilon$, where $X$ is a $5000 \times 10$ matrix and $\beta$ is a $10$ dimensional vector with $3$ true predictors. We assess the aforementioned model selection techniques with four tests, over $10$ trials for the contamination and magnitude tests and over $20$ trials for the coverage test on $1$ and $10$ subsets for the magnitude and coverage tests and $ 1 $ and $ 50 $ subsets for the contamination tests with 1,000 iterations on each MCMC chain and a burn-in period of the initial $500$ iterations.

\begin{figure}[h]
\centering
\includegraphics[width=1.0\linewidth]{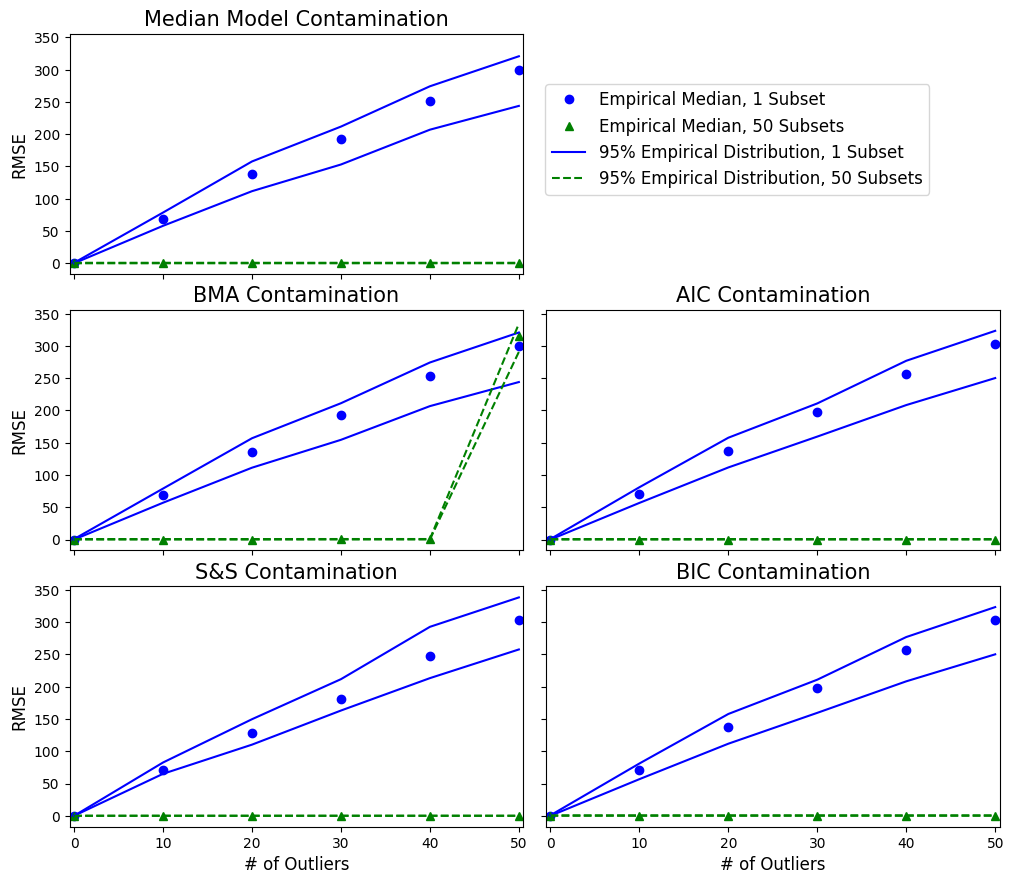}
\caption{Contamination test.}
\label{fig:contamination}
\end{figure}

The first test is the contamination test which examines the root mean square error (RMSE) of held-out test data $\tilde{Y}$ of size $50$ against the number of outliers present (as many as $5$ in our experiments) in the training data, $Y$. We generate outliers by taking the maximum of the absolute value of the data and add a given magnitude value. Each outlier has a relative magnitude of 10,000 meaning that we find the largest output, $ Y_{i^{\ast}}$ such that $ i^{\ast} = \argmax_{i} \left\{ |Y_i|: i = 1, \ldots , N  \right\} $, so that the value of the outlier is $Y_{i^{\ast}} + \left( \mbox{sgn}(Y_{i^{\ast}}) \times 10000  \right)$. For the contamination test, we expect to see superior performance with regards to RMSE of the $50$ subset median posterior as long as the number of outliers per subset does not exceed $1$. Figure~\ref{fig:contamination} demonstrates the robustness of our technique to the number of outliers when we divide the data into subsets. We can see that the empirical $95\%$ distribution of the RMSE over $ 10 $ trials for $50$ subsets (green dashed line) falls dramatically below that of the RMSE distribution of $1$ subset for each model selection technique when outliers are present except in the case when $ 50 $ outliers are present for Bayesian model averaging which approaches the point where the theoretical guarantees of our method are violated.

\begin{figure}[h]
\centering
\includegraphics[width=1.0\linewidth]{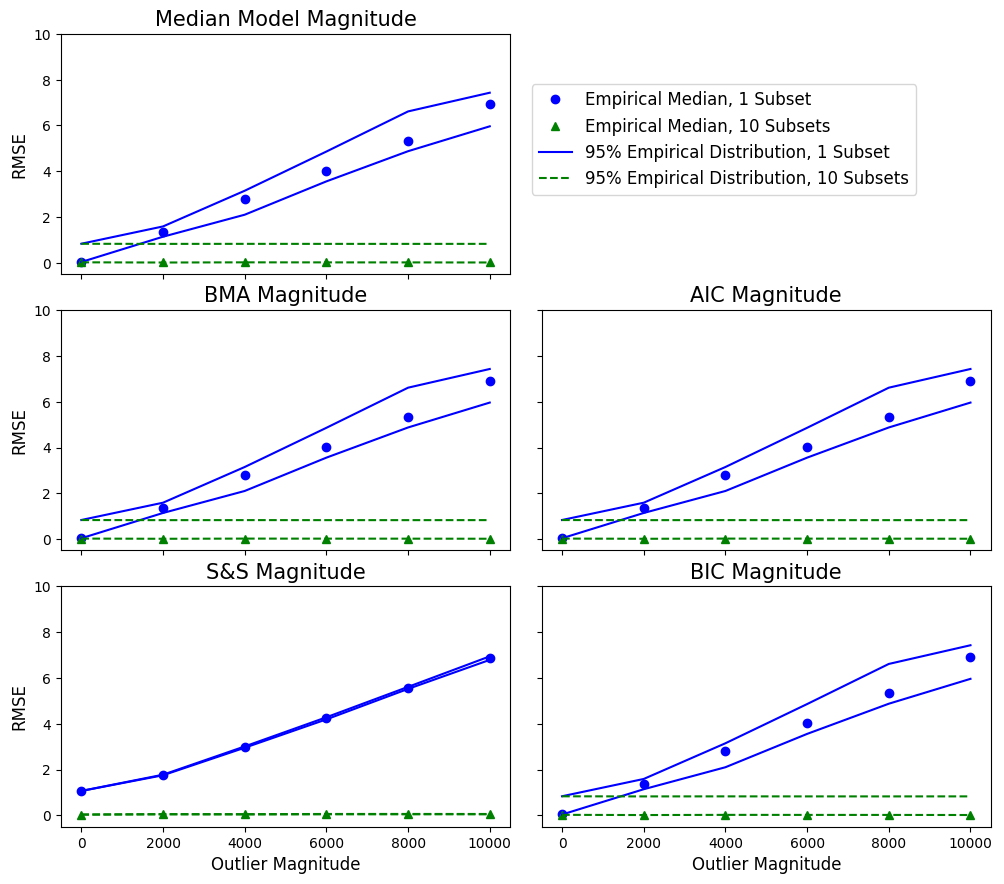}
\caption{Magnitude of outlier test.}
\label{fig:magnitude}
\end{figure}

The second test assesses the RMSE of the held-out test data of size $50$ against the \emph{increasing relative magnitude of one outlier} present in the training data.  We expect to see nearly constant RMSE on the $10$ subset run as the relative magnitude of a single outlier increases, thus the procedure is robust. We can see in Figure~\ref{fig:magnitude} that the RMSE of distributed variants of the model selection techniques are lower than the single processor variants as the number of outliers increases. In the magnitude test, we can categorically observe that $10$ subset RMSE is invariant to the relative magnitude of one outlier present in the data whereas the RMSE grows rapidly on one subset.

\begin{figure}[h]
\centering
\includegraphics[width=1.0\linewidth]{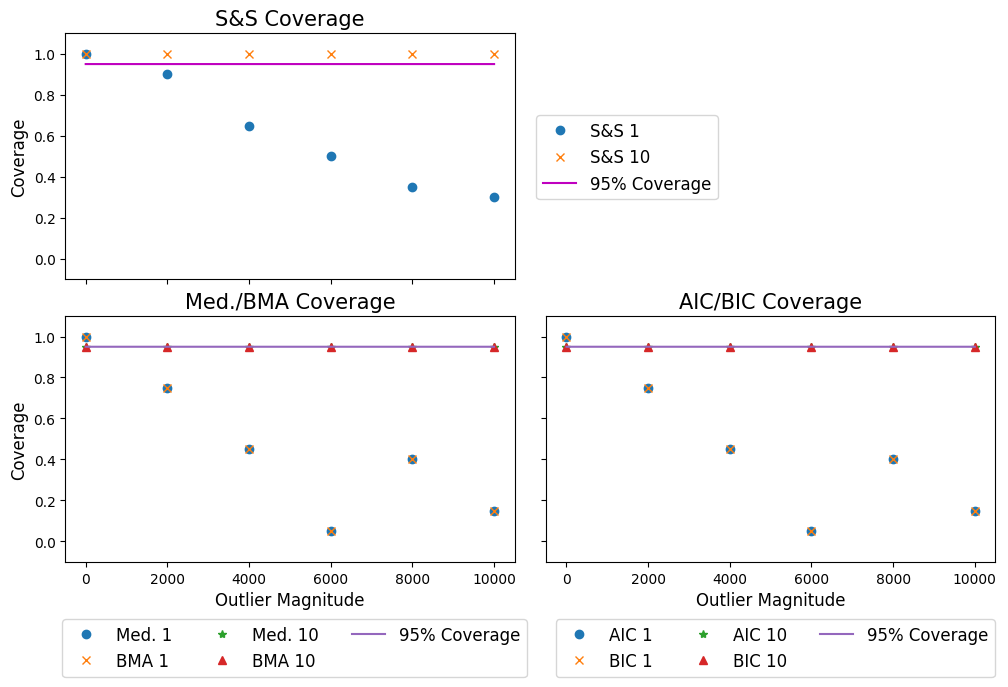}
\caption{Testing empirical coverage of predictive value.}
\label{fig:coverage}
\end{figure}

The next test assesses the $95\%$ frequentist posterior coverage of the true held-out predictive value of size $1$, $\tilde{Y}$, against the \emph{increasing relative magnitude} of one outlier in the training data. To calculate coverage we generate $50$ independent MCMC chains at each level of outlier magnitude and calculate the proportions of chains which include the true predictive value within the $2.5\%$ and $97.5\%$ percentiles of the posterior predictive draws. For the coverage test we see that the empirical coverage of a single predictive value for the distributed subsets is, on average, $95\%$ regardless of the magnitude of the outlier as opposed to the empirical coverage for the single subset. In the $1$ subset case, we can see that the empirical coverage degrades almost to zero as the magnitude of the outlier grows. (see Figure~\ref{fig:coverage}).

\begin{figure}[h]
	\centering
	\includegraphics[width=1.0\linewidth]{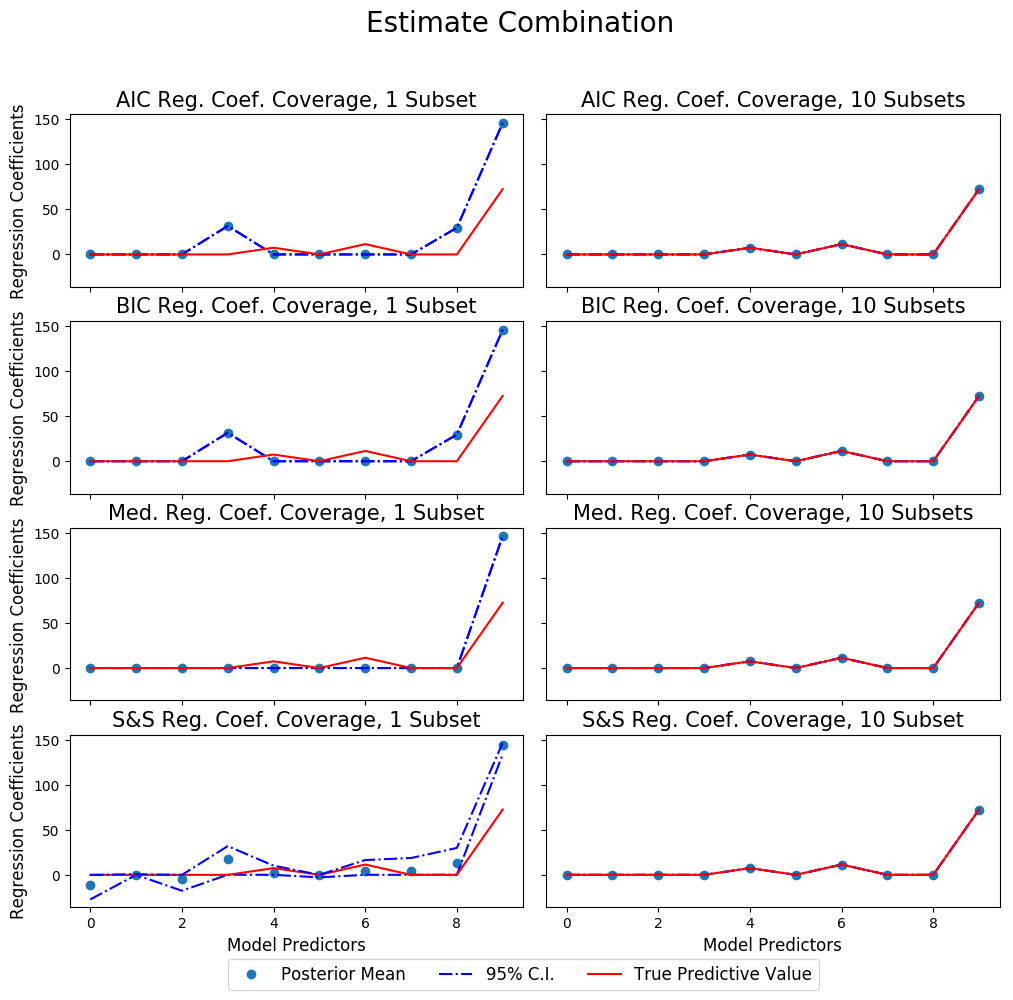}
	\caption{Posterior regression parameter coverage test results, estimate combination.}
	\label{fig:reg_coverage}
\end{figure}

\begin{figure}[h]
	\centering
	\includegraphics[width=1.0\linewidth]{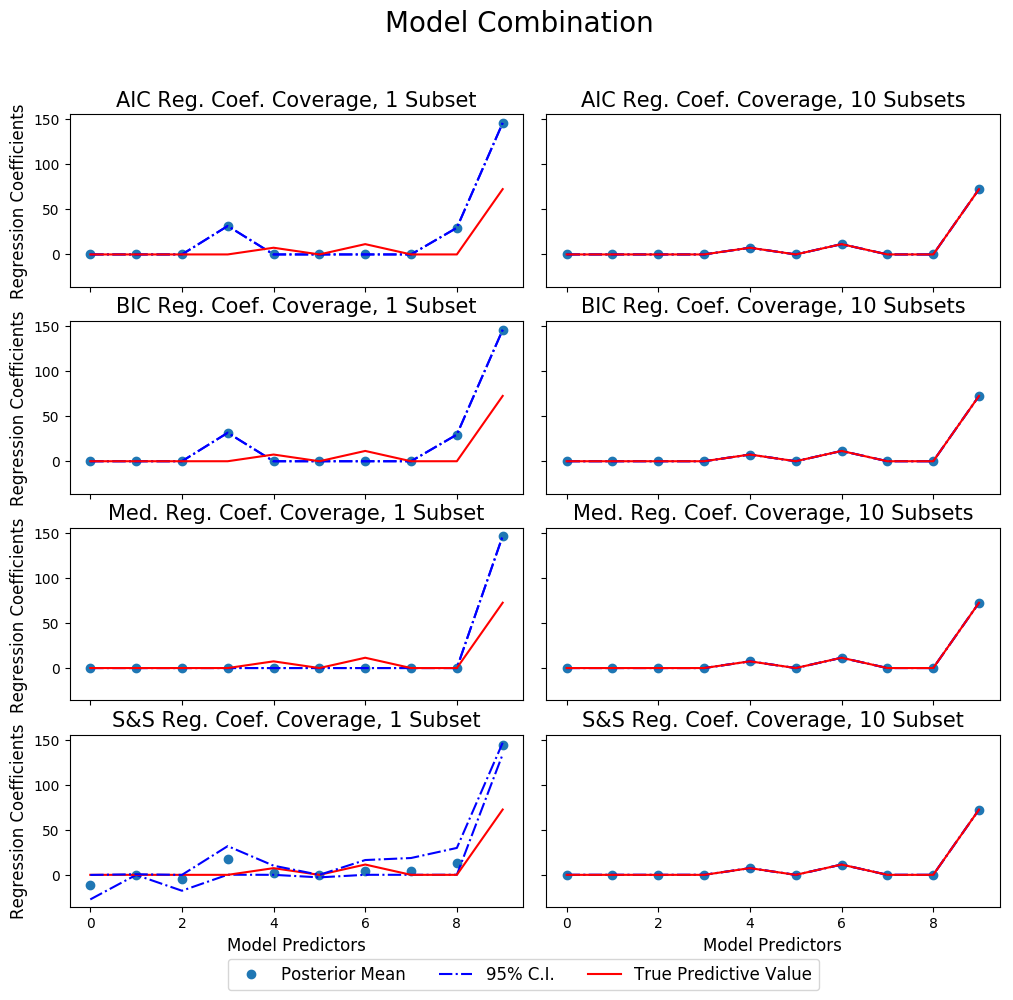}
	\caption{Posterior regression parameter coverage test results, model combination.}
	\label{fig:reg_coverage_model}
\end{figure}

Our last evaluation is the coverage of the regression coefficients and the ability for our model selection techniques to \emph{choose the correct model} under the distributed setting with a single outlier of magnitude 10,000. We compare the posterior credible interval of the regression coefficients for $ 1 $ and $ 10 $ subsets. Note that we do not include nested models in our evaluations or models larger than the true model (i.e models with more than $ 3 $ covariates included). Furthermore, we perform this evaluation under two settings: One, where we combine the optimal local model seleceted on each subset (``Model Combination'') or if we combine the subposterior estimates and select the optimal model globally (``Estimate Combination'') As seen in Figure~\ref{fig:reg_coverage}, the parallel technique is able to select the correct model $ 1 $ subset test, the outlier leads to the incorrect model being selected. Additionally, Figure~\ref{fig:reg_coverage_model} demonstrates that model and estimate combination yield similar results with the regression coefficient coverage test. 

\begin{figure}[hbtp]
	\centering
	\includegraphics[width=0.9\linewidth]{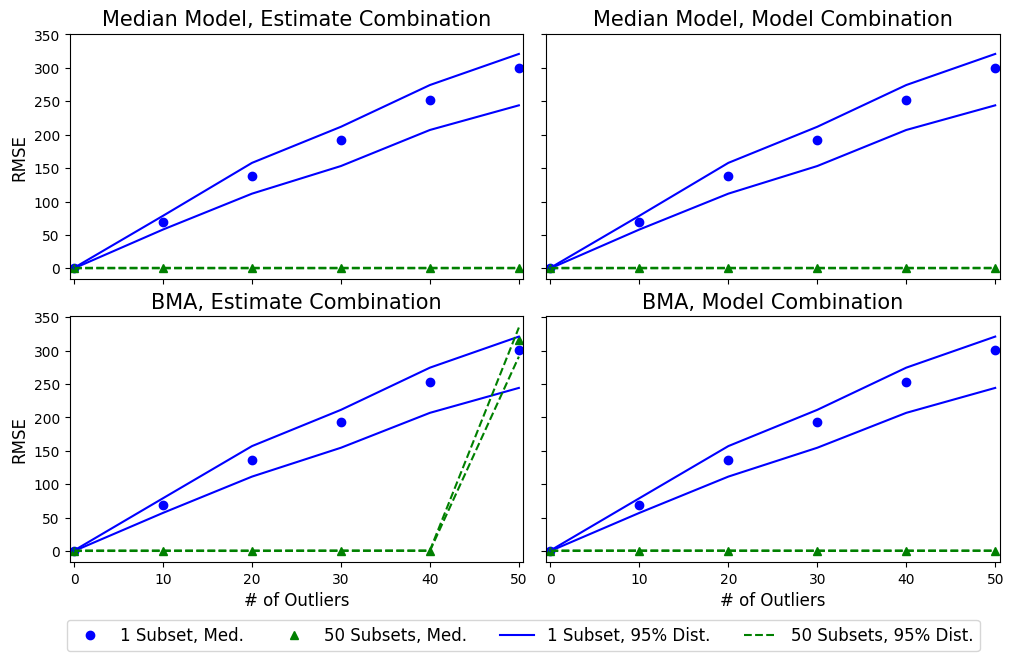}\\	
	\includegraphics[width=0.9\linewidth]{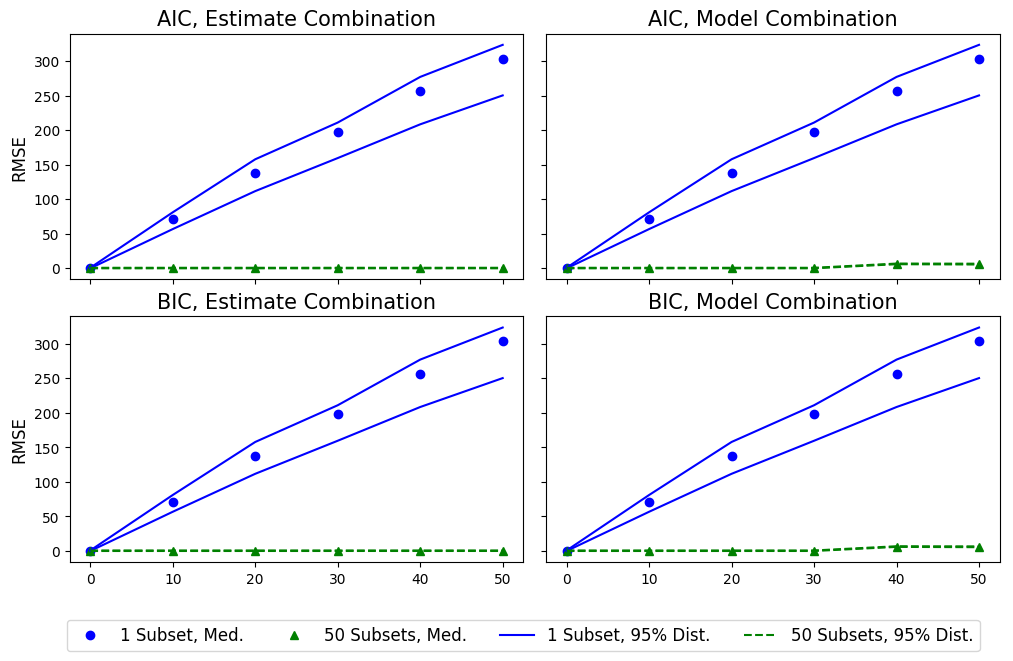}\\
	\includegraphics[width=0.9\linewidth]{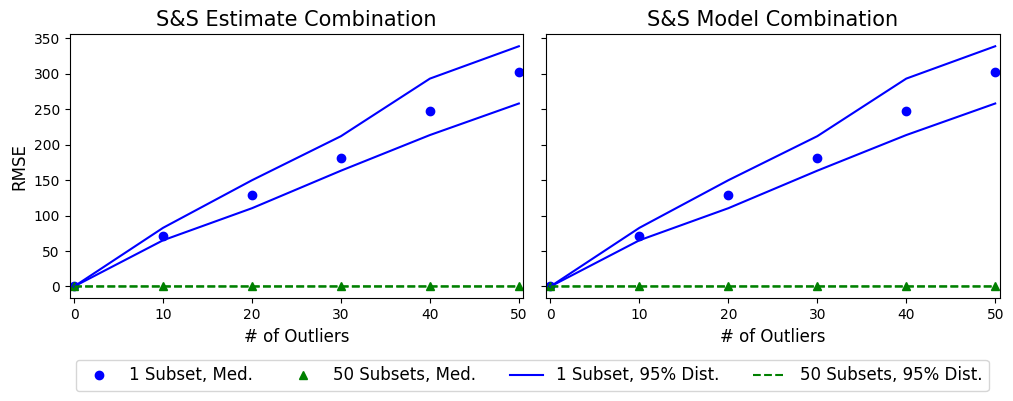}	
	\caption{Contamination test}\label{fig:contam_compare}
\end{figure}

\begin{figure}[hbtp]
	\centering
	\includegraphics[width=0.9\linewidth]{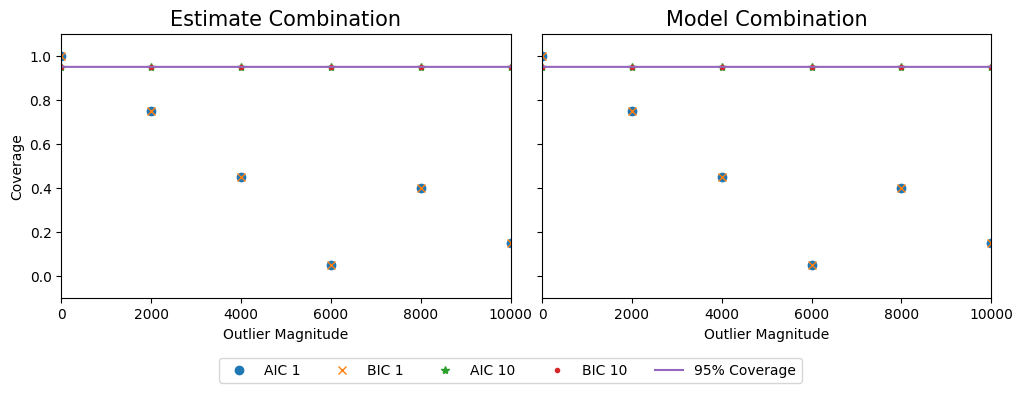}\\
	\includegraphics[width=0.9\linewidth]{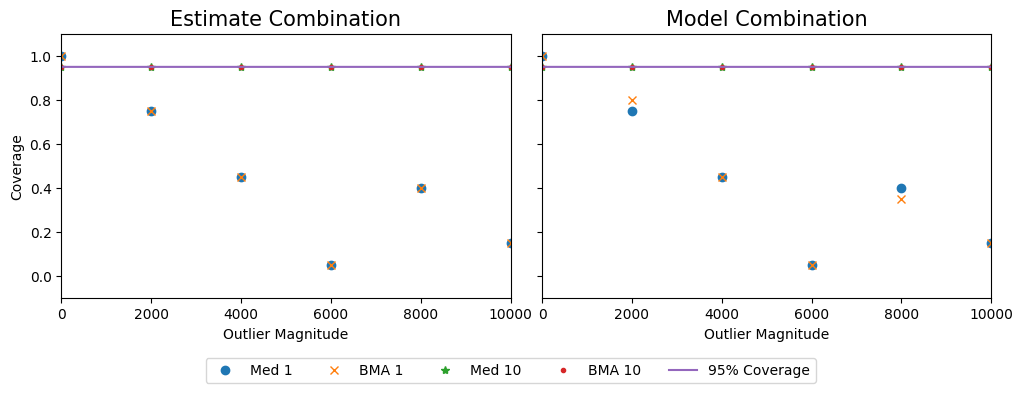}
	\includegraphics[width=0.9\linewidth]{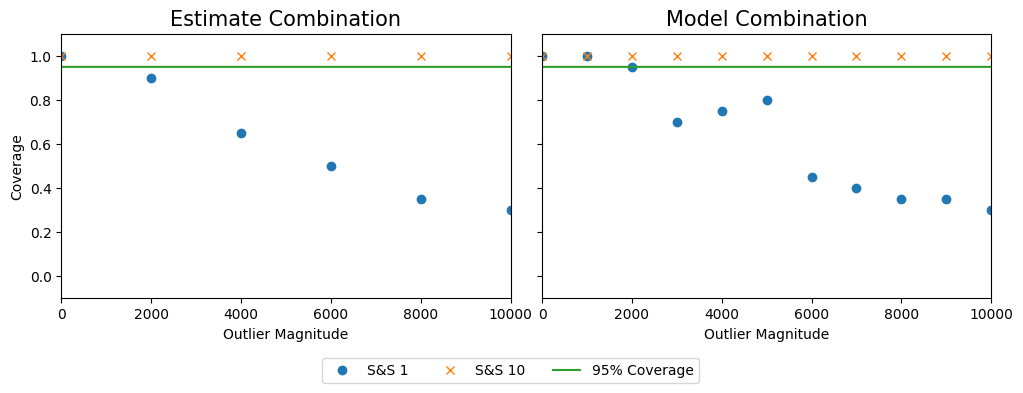}	
	\caption{Coverage test.}\label{fig:coverage_compare}
\end{figure}

\begin{figure}[hbtp]
	\centering
	\includegraphics[width=0.9\linewidth]{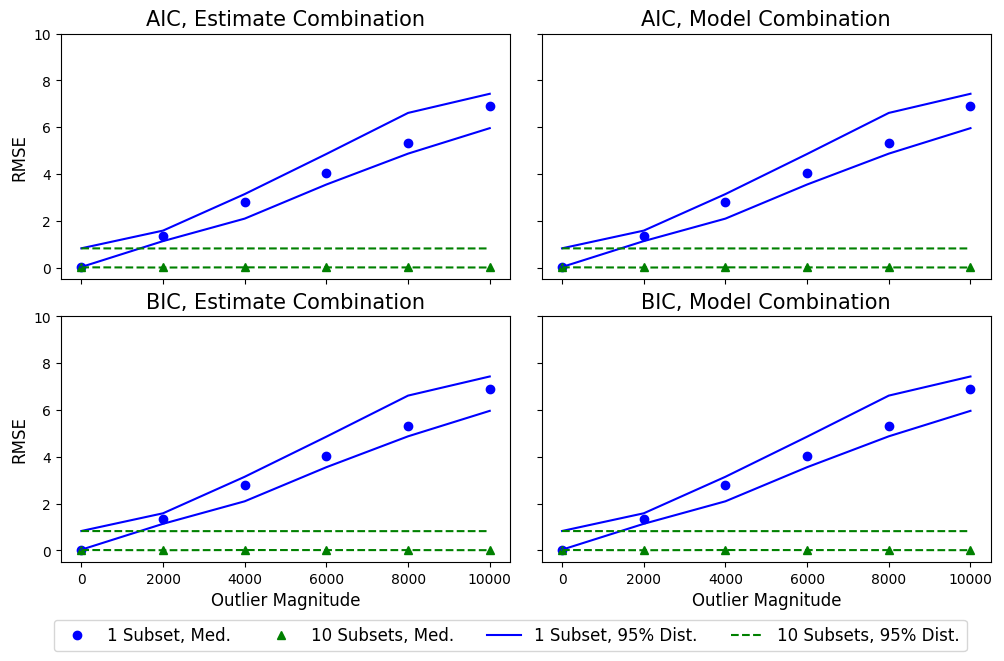}\\
	\includegraphics[width=0.9\linewidth]{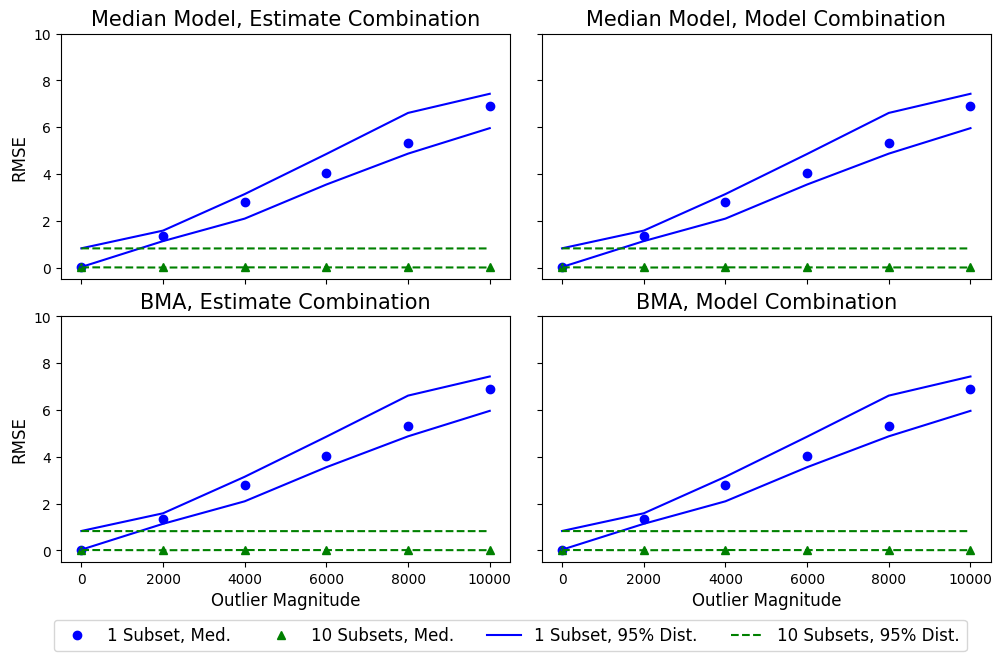}\\
	\includegraphics[width=0.9\linewidth]{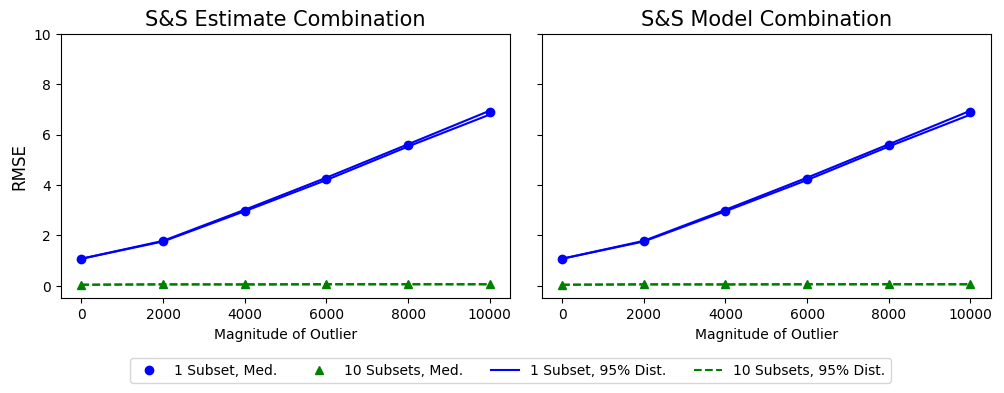}
	\caption{Magnitude test.}\label{fig:magnitude_compare}
\end{figure}

Also, we would like to see if the results still hold between model and estimate combination for the other simulation studies performed. Figs.~\ref{fig:contam_compare}, \ref{fig:coverage_compare}, and \ref{fig:magnitude_compare} show that there is little difference in how we combine the information for model selection in each of the tests evaluated. 

\begin{figure}[hbtp]
	\centering
	\includegraphics[width=0.9\linewidth]{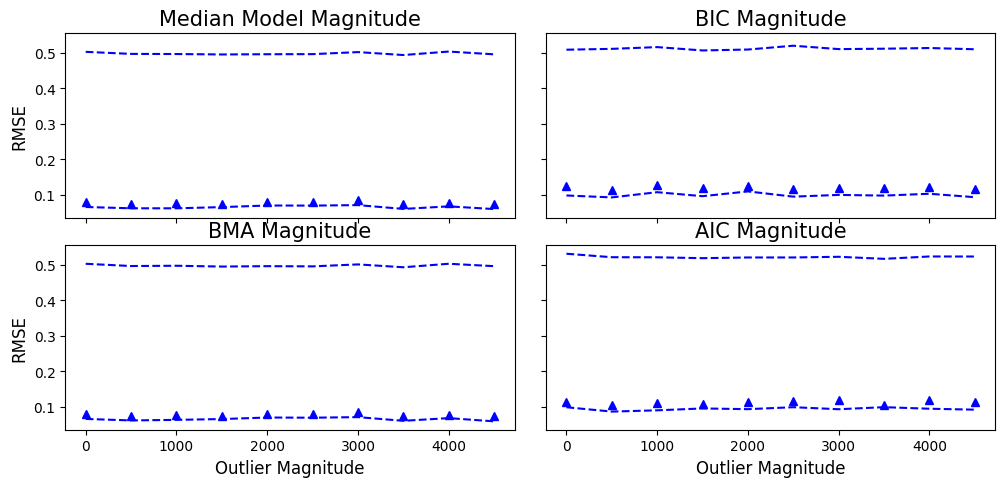}\\
	\includegraphics[width=0.9\linewidth]{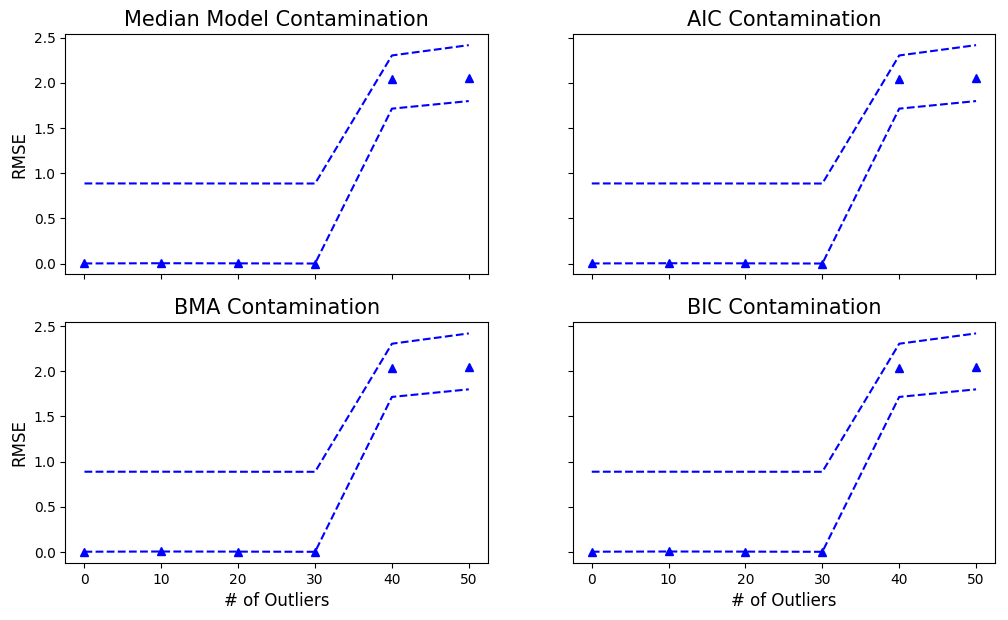}\\
	\caption{Synthetic big data results.}\label{fig:big_data}
\end{figure}

Furthermore, we wish to evaluate our method a large synthetic dataset with the same synthetic generating process as above, but with \emph{one million observations} divided over 50 processors. Here, we examine the behavior of our method when we increase the magnitude of one outlier in the dataset and when we increase the number of outliers with fixed magnitude. In Figure~\ref{fig:big_data}, we can see that our performance is robust when the number of outliers per subset fulfills Theorem 4. When the number of outliers reaches 40 and 50, we see start to see a noticeable degradation of our method's predictive ability. However, this degradation is still small relative to what we might observe in the case where we do not divide the data into subsets. 

Additionally, we would like to see the computational gain of dividing the data for this situation in terms of CPU time for running the model selection and inference procedure. For one subset the average computation time is 91,829.15  seconds with a standard error of $ 190.80 $ seconds.  For ten subsets, the average computation time is 10,301.60  seconds with a standard error of $ 81.28 $ seconds. And for fifty subsets, the average computation time is  29,49.74  seconds with a standard error of $ 16.61 $ seconds which signifies that we obtain critical computational performance when dividing our method across multiple processors.

\begin{figure}[hbtp]
\centering
\includegraphics[width=0.9\linewidth]{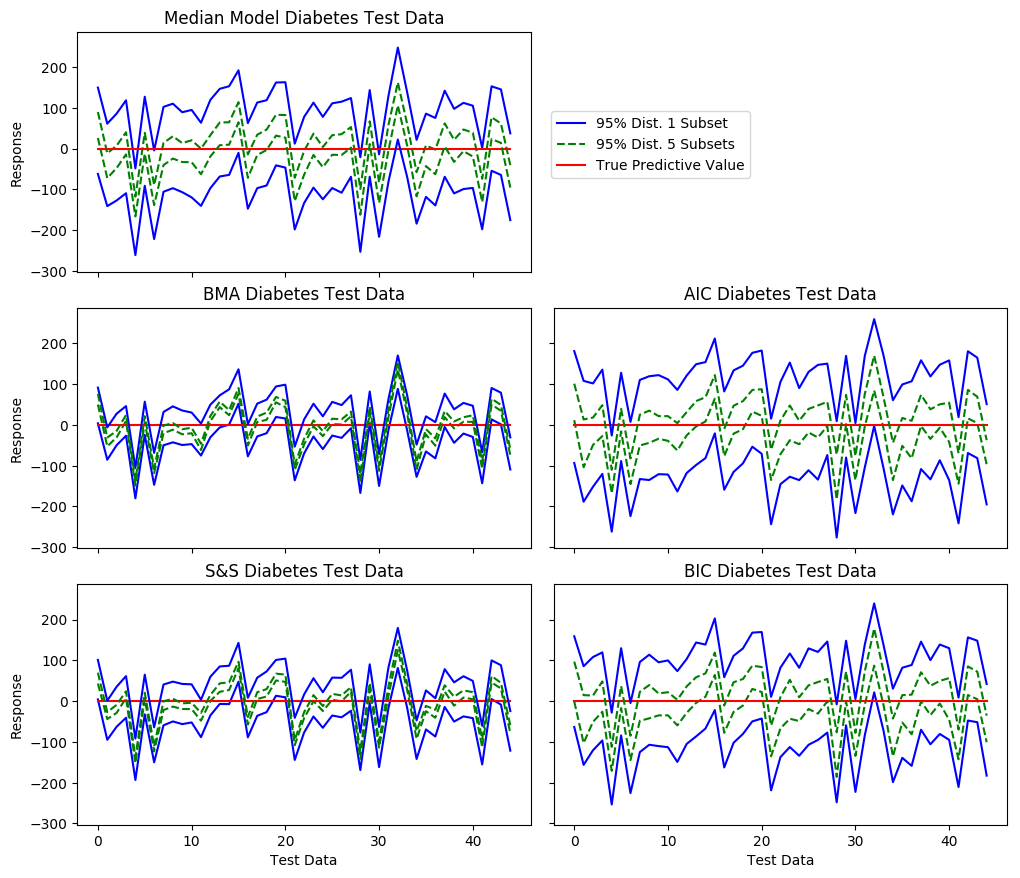}
\caption{Diabetes test data results.}\label{fig:diabetes}
\end{figure}

Lastly, we evaluate our parallel model selection method on the diabetes data set used in \cite{efron2004least}. The diabetes data consists of a $442 \times 10$ dimension design matrix scaled with unit norm and zero mean and a single response vector. We held out $45$ observations for test evaluation and plotted the posterior $95\%$ credible intervals for the predictive values centered at zero after subtracting the true predictive value. We can see in Fig.~\ref{fig:diabetes} that, after dividing the data across 5 subsets, we can attain a tighter credible interval over the true value for each model selection technique.



	\section{CONCLUSION}\label{sec:conclusion}
While a substantial body of work exists for fast and scalable Bayesian inference methods, few research methods are available on robust and scalable model selection.
We have studied in this paper a divide-and-conquer strategy that contributes to filling this gap. This strategy operates by taking the geometric median of posterior model probabilities or other selection criteria that extends previous results focusing on parametric inference. We show theoretically how the strategy, particularly in the setting of BMA, can be robust to outliers and, moreover, exhibits faster concentration to the true model in terms of posterior model probabilities. The concentration result also applies to the joint setting of model selection and parameter estimation. We illustrate with both simulation data and a real data example how a variety of our strategy leads to more robust inference compared to standard approach that does not divide data into subsets. The strategy we present is simple to execute and is foreseen to have good practical value.

%
%


	\section{APPENDIX}\label{sec:appendix}
\noindent\textbf{Proof of Theorem \ref{thm plain}.}

The proof is a modification of that for Theorem 2.1 in \cite{ghosal2000}. Take any $\epsilon>2\varepsilon_N$, we have, by Assumption \ref{assumption1},
$$\log\mathcal D\left(\frac{\epsilon}{2},\mathcal P_{\mathcal S_N},d\right)\leq\log\mathcal D\left(\varepsilon_N,\mathcal P_{\mathcal S_N},d\right)\leq N\varepsilon_N^2.$$
Then, by Theorem 7.1 in \cite{ghosal2000}, there exists tests $\phi_N$ and a large enough constant $T$ (chosen later) such that
\begin{equation}
P_0^N\phi_N\leq e^{N\varepsilon_N^2}e^{-LNT^2\varepsilon_N^2}\frac{1}{1-e^{-LNT^2\varepsilon_N^2}},\label{test1}
\end{equation}
and
\begin{equation}
\sup_{\theta\in\mathcal S_N:d(P_\theta,P_0)>T\varepsilon_N}P_\theta^N(1-\phi_N)\leq e^{-LNT^2\varepsilon_N^2},\label{test2}
\end{equation}
for a universal constant $L>0$, any $N>0$, and $P_\theta^N$ denotes the probability measure on $(X,Y)$ under $(X_1,Y_1)\sim p_0(x)\times p_\theta(y|x)$.

By \eqref{test1}, we have
\begin{equation}
P_0^NPr(\theta:\theta:d(P_\theta,P_0)>L\varepsilon_N^2|X,Y)\phi_N\leq P_0^N\phi_N\leq2e^{-LN\varepsilon_N^2},\label{interim5}
\end{equation}
as $N\to\infty$, if we choose $LT^2-1>L$. Now, since
$$P_0\frac{p_\theta(Y_1|X_1)}{p_0(Y|X)}=\int\frac{p_\theta(y|x)}{p_0(y|x)}p_0(dy|x)p_0(dx)=\int p_\theta(dy|x)p_0(dx)=1,$$
by Fubini's theorem, we have
$$P_0^N\int_{\mathcal S\setminus\mathcal S_N}\prod_{i=1}^N\frac{p_\theta(Y_i|X_i)}{p_0(Y_i|X_i)}Pr(\mathrm{d}\theta)\leq Pr(\mathcal S\setminus\mathcal S_N).$$
Hence, by Fubini's theorem again,
\begin{eqnarray}
&&P_0^N\int_{\theta\in\mathcal S:d(P_\theta,P_0)>T\varepsilon_N}\prod_{i=1}^N\frac{p_\theta(Y_i|X_i)}{p_0(Y_i|X_i)}Pr(\mathrm{d}\theta)(1-\phi_N)\notag\\
&\leq&\Pi(\mathcal S\setminus\mathcal S_N)+\int_{\theta\in\mathcal S_N:d(P_\theta,P_0)>T\varepsilon_N}P_\theta^N(1-\phi_N)Pr(\mathrm{d}\theta)\notag\\
&\leq&\Pi(\mathcal S\setminus\mathcal S_N)+e^{-LNT^2\varepsilon_N^2}\text{\ \ by \eqref{test2}}\notag\\
&\leq&2e^{-N\varepsilon_N^2(C+4)},\label{interim4}
\end{eqnarray}
if $KM^2\geq C+4$, by Assumption \ref{assumption2}.

By Lemma \ref{lemma bound} (stated below) and Assumption \ref{assumption3}, with probability at least $1-1/(C^2N\varepsilon_N^2)$, we have
\begin{equation}
\int\prod_{i=1}^N\frac{p_\theta(Y_i|X_i)}{p_0(Y_i|X_i)}Pr(\mathrm{d}\theta)\geq e^{-2N\varepsilon_N^2}Pr(B_n)\geq e^{-N\varepsilon_N^2(2+C)},\label{interim3}
\end{equation}
where
$$B_n=\left\{\theta:-P_0\log\frac{p_\theta(Y_1|X_1)}{p_0(Y_1|X_1)}\leq\varepsilon_N^2,\ P_0\left(\frac{p_\theta(Y_1|X_1)}{p_0(Y_1|X_1)}\right)^2\leq\varepsilon_N^2\right\}.$$
Let $A_N$ be the event that \eqref{interim3} holds. We have
\begin{eqnarray*}
	&&P_0^NPr(\theta:d(P_\theta,P_0)>T\varepsilon_N|X,Y)(1-\phi_N)\mathbf 1_{A_N}\\
	&=&P_0^N\frac{\int_{\theta:d(P_\theta,P_0)>T\varepsilon_N}\prod_{i=1}^N\frac{p_\theta(Y_i|X_i)}{p_0(Y_i|X_i)}Pr(\mathrm{d}\theta)}{\int\prod_{i=1}^N\frac{p_\theta(Y_i|X_i)}{p_0(Y_i|X_i)}Pr(\mathrm{d}\theta)}(1-\phi_N)\mathbf 1_{A_N}\\
	&\leq&e^{N\varepsilon_N^2(2+C)}2e^{-N\varepsilon_N^2(C+4)}\text{\ \ by \eqref{interim4} and \eqref{interim3}}\\
	&=&2e^{-2N\varepsilon_N^2}.\label{interim6}
\end{eqnarray*}

Therefore,
\begin{eqnarray*}
	&&P_0^NPr(\theta:d(P_\theta,P_0)>T\varepsilon_N|X,Y)\\
	&=&P_0^NPr(\theta:d(P_\theta,P_0)>T\varepsilon_N|X,Y)\phi_N+P_0^NPr(\theta:d(P_\theta,P_0){}\\
	&&{}>T\varepsilon_N|X,Y)(1-\phi_N)\mathbf 1_{A_N}+P_0^NPr(\theta:d(P_\theta,P_0){}\\
	&&>{}T\varepsilon_N|X,Y)(1-\phi_N)(1-\mathbf 1_{A_N})\\
	&\leq&P_0^NPr(\theta:d(P_\theta,P_0)>T\varepsilon_N|X,Y)\phi_N+P_0^NPr(\theta:d(P_\theta,P_0){}\\
	&&{}>T\varepsilon_N|X,Y)(1-\phi_N)\mathbf 1_{A_N}+P_0^N(A_N^c)\text{\ \ for sufficiently large $T$}\\
	&\leq&2e^{-LN\varepsilon_N^2}+2e^{-2N\varepsilon_N^2}+\frac{1}{C^2N\varepsilon_N^2},
\end{eqnarray*}
by \eqref{interim5}, \eqref{interim6} and the property of $A_N$. By Chebyshev's inequality, we have
$$P_0^N\left(Pr(\theta:d(P_\theta,P_0)>T\varepsilon_N^2|X,Y)>\delta\right)\leq\frac{1}{C^2N\varepsilon_N^2\delta}+\frac{2e^{-LN\varepsilon_N^2}}{\delta}+\frac{2e^{-2N\varepsilon_N^2}}{\delta},$$
which concludes the theorem.\hfill\qed

\begin{lemma}
	For any $\epsilon>0$ and probability distribution $\Pi$ defined on the set
	\begin{equation}
	\left\{\theta:-P_0\log\frac{p_\theta(Y|X)}{p_0(Y_1|X_1)}\leq\epsilon^2,\ P_0\left(\frac{p_\theta(Y|X)}{p_0(Y_1|X_1)}\right)^2\leq\epsilon^2\right\},\label{set}
	\end{equation}
	we have, for every $C>0$,
	\begin{equation}
	P_0^N\left(\int\prod_{i=1}^N\frac{p_\theta(Y_i|X_i)}{p_0(Y_i|X_i)}\Pi(\mathrm d\theta)\leq e^{-(1+C)N\epsilon^2}\right)\leq\frac{1}{C^2N\epsilon^2}.\label{bound}
	\end{equation}\label{lemma bound}
\end{lemma}




\begin{thm}[Adopted from \cite{minsker2015geometric}]
	Consider a Hilbert space $(\mathbb H,\langle\cdot,\cdot\rangle)$ and $\xi_0\in\mathbb H$. Let $\hat\xi_1,\ldots,\hat\xi_R\in\mathbb H$ be a collection of independent random $\mathbb H$-valued elements. Let  $\alpha,q,\nu$ be constants such that $0<q<\alpha<1/2$ and $0\leq\nu<(\alpha-q)/(1-q)$. Suppose that there exists $\epsilon>0$ such that for all $j$, where $1\leq j\leq\lfloor(1-\nu)R\rfloor+1$,
	$$P(\|\hat\xi_j-\xi_0\|>\epsilon)\leq q.$$
	Let $\hat\xi_*=\text{med}_g(\hat\xi_1,\ldots,\hat\xi_R)$ be the geometric median of $\{\hat\xi_1,\ldots,\hat\xi_R\}$. Then
	$$P(\|\hat\xi_*-\xi_0\|>C_\alpha\epsilon)\leq\left(e^{(1-\nu)\psi(\frac{\alpha-\nu}{1-\nu},q)}\right)^{-R},$$
	where $C_\alpha=(1-\alpha)\sqrt{1/(1-2\alpha)}$, and
	$$\psi(\alpha,q)=(1-\alpha)\log\frac{1-\alpha}{1-q}+\alpha\log\frac{\alpha}{q}.$$\label{thm geometric median}
\end{thm}

\section*{ACKNOWLEDGMENTS}
The contribution of Lizhen Lin was funded by NSF grants IIS 1663870, CAREER DMS 1654579. The contribution of Henry Lam was funded by NSF grants CMMI-1542020, CMMI-1523453 and CAREER CMMI-1653339. The contribution of Michael Zhang was funded by NSF grant 1447721. 

\section*{REFERENCES}
	\bibliography{robust_bayes}
	
\end{document}